\pdfoutput=1

\documentclass[11pt]{article}

\usepackage{acl}

\usepackage{times}
\usepackage{latexsym}

\usepackage{makecell}

\usepackage[T1]{fontenc}

\usepackage[utf8]{inputenc}

\usepackage{microtype}

\usepackage{inconsolata}

\usepackage{graphicx}

\usepackage{amssymb,amsmath}
\usepackage{makecell}

\usepackage{booktabs} 
\usepackage{multirow} 

\title{MLKV: Multi-Layer Key-Value Heads for Memory Efficient \\Transformer Decoding}

\author{Zayd Muhammad Kawakibi Zuhri$^1$, Muhammad Farid Adilazuarda$^2$, \\ \textbf{Ayu Purwarianti$^1$,} \textbf{Alham Fikri Aji}$^2$ \\
  $^1$Institut Teknologi Bandung \quad $^2$MBZUAI \\
  \texttt{zaydzuhri@gmail.com}}

\begin{document}
\maketitle
\begin{abstract}
Auto-regressive inference of transformers benefit greatly from Key-Value (KV) caching, but can lead to major memory bottlenecks as model size, batch size, and sequence length grow at scale. We introduce Multi-Layer Key-Value (MLKV) sharing, a novel approach extending KV sharing across transformer layers to reduce memory usage beyond what was possible with Multi-Query Attention (MQA) and Grouped-Query Attention (GQA). Evaluations on various NLP benchmarks and inference metrics using uptrained Pythia-160M variants demonstrate that MLKV significantly reduces memory usage with minimal performance loss, reducing KV cache size down to a factor of 6x compared to MQA. These results highlight MLKV's potential for efficient deployment of transformer models at scale\footnote{We will release all the code implementation and experimental results of our proposed architecture}. 
\end{abstract}

\section{Introduction}
The transformer architecture \citep{vaswani2017attention} has brought about Large Language Models (LLMs) that excel in natural language processing. However, due to its auto-regressive nature when doing inference, the decoder is bottlenecked by memory bandwidth when storing and loading keys and values at each time-step, also called KV caching. Because this cache scales linearly with model size, batch size, and context length, it can even exceed the memory usage of the model weights themselves \citep{pope2022efficiently}.

The most notable methods for handling large KV caches are the approaches that directly reduce the number of KV heads used, from here on referred to as KV sharing. Multi-Query Attention (MQA) by \citet{shazeer2019fast} uses only a single key and value projection for all attention heads in a layer. It reduces memory bandwidth for KV cache by $1/n\_heads$, which is significant, but results in some degradation in quality and stability. Grouped-Query Attention (GQA) from \citet{ainslie2023gqa} addresses these issues and introduces an intermediate solution of sharing several KV heads to multiple groups of attention heads. This effectively reduces KV cache size by $n\_groups/n\_heads$. They also introduce an uptraining scheme to convert any transformer checkpoint to MQA or GQA. With a reasonable number of heads, GQA can achieve near parity on benchmarks with a vanilla model. However, the reduction in KV cache size is still limited to $1/n\_heads$ with MQA, which still might not be enough for some applications.

\begin{figure}[t]
    \centering
    \includegraphics[width=\columnwidth]{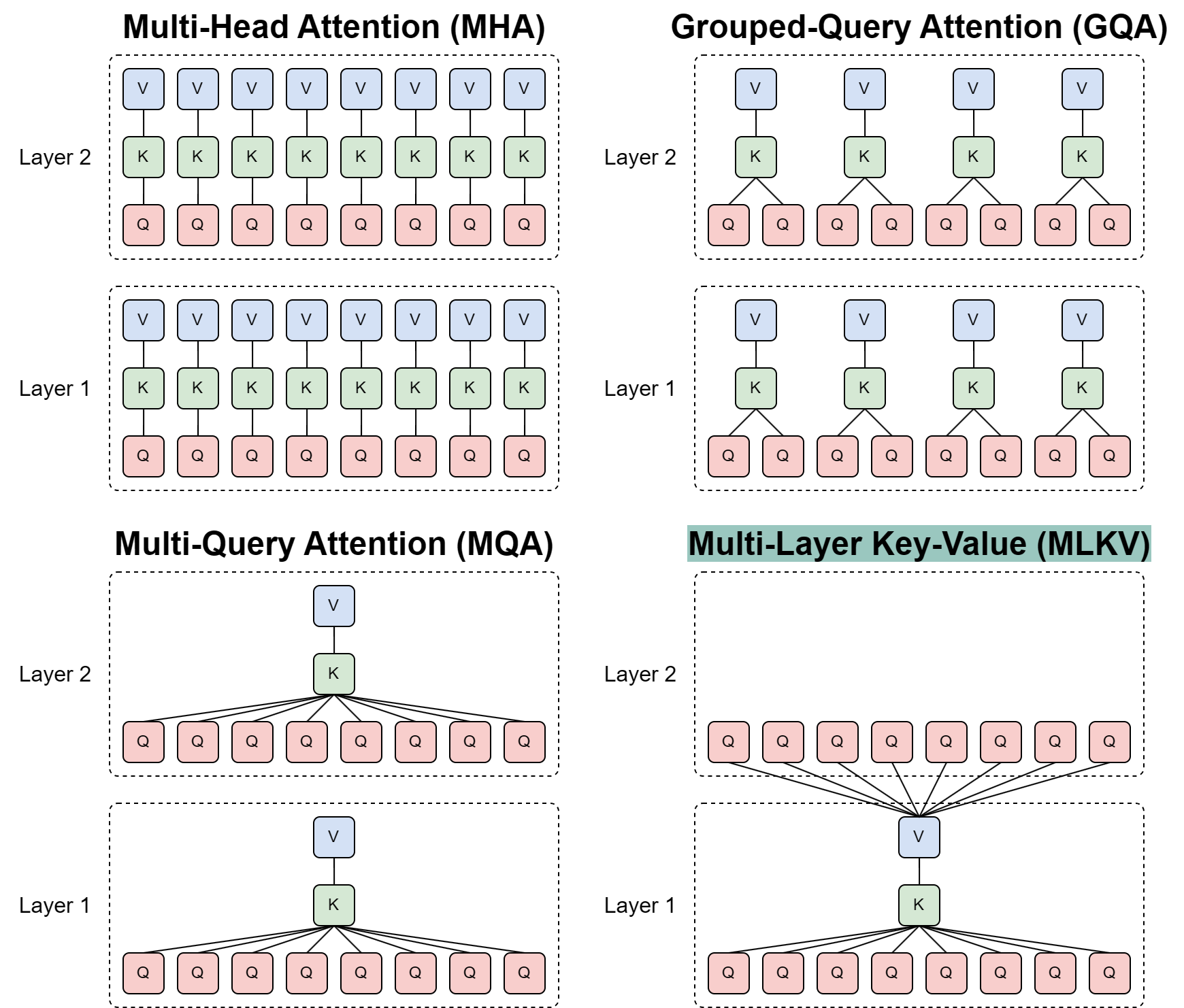}
    \caption{Simplified overview of current KV sharing methods, vanilla MHA (top left), MQA (bottom left), and GQA (top right). All of them share KV heads within the same layer. Our proposed KV sharing scheme MLKV (bottom right) shares KV heads between layers.}
    \label{fig:mqagqaviz}
\end{figure}

To go beyond this limitation, we introduce Multi-Layer Key-Value (MLKV) sharing. Taking KV sharing one step further, MLKV not only shares KV heads among attention heads in the same layer, but also among heads in other layers. KV heads can be used on groups of heads in the same layer and groups of heads in the next layers too. At the most extreme, a single KV head can be used for all heads in all layers. We adopt the uptraining strategy of GQA to MLKV and uptrain from Pythia-160M checkpoints. We experiment with configurations that utilize both grouped queries in the same layer and among different layers, but also MLKV-only configurations that go beyond MQA, with KV head counts lower than the number of layers. We show that these configurations provide a reasonable performance trade-off for the memory savings achieved, up to $2/n\_layers$ the original KV cache size without a significant model degradation.


\begin{figure*}[t]
    \includegraphics[width=\linewidth]{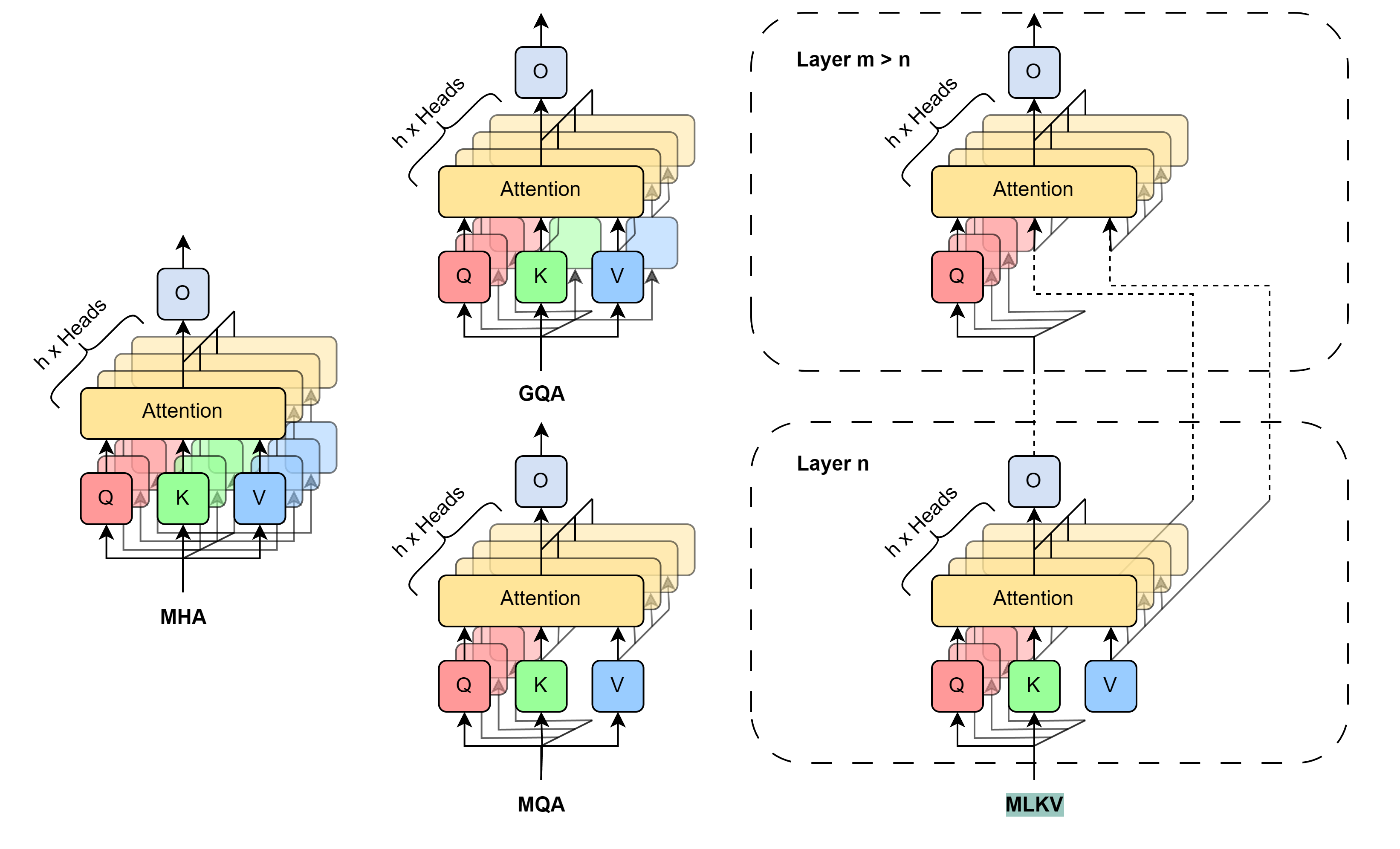}
    \caption{Detailed illustration of attention using the different KV sharing mechanisms. Vanilla MHA (left) has a key-value head for each query head. GQA (top middle) here with 2 groups of heads. MQA (bottom middle) only has one key-value head for all query heads. MLKV (right) can share the one key-value head from the bottom layer, to the query heads of some layer above it.}
    \label{fig:allkvmechs}
\end{figure*}

\section{Background}
\label{background}
\subsection{Multi-Head Attention (MHA)}
The vanilla transformer introduced by \citet{vaswani2017attention} uses an attention mechanism with multiple "heads". A head is a linear down-projection from the model dimension that is expected to each attend to different representation subspaces. Given $h$ heads, each $i$-th head has a query, key, and value projection with weights $W_i^Q \in \mathbb{R}^{d \times d_q}, W_i^K \in \mathbb{R}^{d \times d_k}, W_i^V \in \mathbb{R}^{d \times d_v}$ that project from the model dimension $d$ to a smaller size which is usually set to $d_q=d_k=d_v=d/h$. In MHA, each query head has its own Key-Value (KV) head with unique weights. The $l$ layers in a transformer receive a sequence of $s$ embedded tokens $x \in \mathbb{R}^{s \times d}$ and are defined as follows (normalization omitted for conciseness, concatenation of the last dimension of $n$ tensors written as $[x_1;...;x_n]$):
\begin{gather}
    q_i = xW_i^Q;\; k_i = xW_i^K;\; v_i = xW_i^V \\
    \alpha_i = softmax(\frac{q_i k_i^T}{\sqrt{d_k}}) \\
    o_i = Attention(q_i,k_i,v_i) = \alpha_i v_i \\
    MHA(x) = [o_1;...;o_h]W^O \label{mha} \\
    L_n(x) = x + MLP(x + MHA(x)) \\
    Layers(x) = L_1 \circ \cdot \cdot \cdot \circ L_l(x)
\end{gather}

\subsection{Multi-Query Attention (MQA)}
\citet{shazeer2019fast} showed that it is possible and viable to share a single KV head for all query heads in the attention of a layer.
\begin{gather}
    q_i = xW_i^Q;\; k = xW^K;\; v = xW^V \\
    o_i = Attention(q_i, k, v) \\
    MQA(x) = [o_1;...;o_h]W^O \label{mqa}
\end{gather}

\subsection{Grouped-Query Attention (GQA)}
\citet{ainslie2023gqa} then proposed to generalize the KV sharing to groups of query heads. This allows for a more flexible configuration and a better performance trade-off. Given $g$ groups of query heads each having one KV head, GQA is defined as follows with $g<h$:
\begin{gather}
    q_i = xW_i^Q;\; k_j = xW_j^K;\; v_j = xW_j^V \\
    o_{i,j} = Attention(q_i, k_j, v_j) \\
    \begin{gathered}
        GQA(x) = [o_{1,1};o_{i,j};...;o_{h,g}]W^O \label{gqa} \\    
        with\; j = floor(\frac{i-1}{h/g}) + 1
    \end{gathered}
\end{gather}

\subsection{KV Caching}
\label{cache}
The reason why these KV sharing methods were developed was to reduce the memory overhead of the KV cache. KV caching is used to optimize autoregressive inference by only keeping the key and value activations of previous tokens. This way, only the singular newest token needs to be passed into the model to generate the next token in the sequence. Given batch size $b$ and sequence length $s$, the KV cache has the dimensions $[2, b, s, l, h, d_k]$. Thus, it scales linearly with batch size, sequence length, and model size, which means it can grow indefinitely during inference. Taking OPT-175B \citep{zhang2022opt} as an example, its parameters require 325GB of memory. Yet when inferencing a sequence length of 2048 and computing a batch of 128 generations at once, the KV cache can take up to 950GB of memory \citep{liu2023scissorhands}.



\subsection{Current Limitations}
\label{currentlimits}
MQA and GQA both provide useful trade-offs in memory overhead and performance over MHA, but are fundamentally limited in how much memory they can save. With MHA, the KV cache has the size $2bslhd_k$ following its dimensions. GQA reduces this to $2bslgd_k$ with $g<h$, meanwhile MQA goes down to $2bsld_k$. At most, the smallest number of KV heads one can achieve is using MQA, where $h = 1$ and the total number of KV heads $mg=l$ as the KV heads are only shared between heads in the same layer. Every layer still must have one KV head, and thus the KV cache can only be reduced at most to $\frac{1}{h}$ of its original size. There is still room to improve here to further decrease memory usage.

\section{Multi-Layer Key-Value (MLKV)}
\label{method}
Since the $h$ dimension of the KV cache has been tackled by MQA and GQA by reducing the number of KV heads in a layer, the next logical step would be to expand sharing to the $l$ dimension, which is the number of layers. This expansion can be grounded by the recent exploration in the role of feed-forward layers in the computation of transformers, mainly by \citet{geva-etal-2021-transformer}. They propose that the feed-forward neural networks in transformer layers emulate key-value memories that process different levels of information. Most notably though are their findings which indicate that groups of successive layers compute similar things. More specifically, lower layers attend to shallow patterns and upper layers to more semantic ones. Thus, it can be also inferred that attention can be delegated to groups of layers, while retaining the needed computation in the feed-forward networks. Intuitively, KV heads can be shared among layers that are assumed to have similar purposes.

Expanding upon those ideas, we propose Multi-Layer Key-Value sharing or MLKV. MLKV not only shares KV heads among query heads in the same layer, like in MQA or GQA, but also among heads in other layers. This allows the total number of KV heads in the transformer to go below what is possible with MQA, and thus allowing for an even smaller KV cache. With $m$ being the number of layers that have their own KV heads, the KV cache has the size $2bsmgd_k$ using MLKV. If we set $g=1$ (like MQA) and $m<l$, this allows for a KV cache that is $\frac{m}{l}$ the original size. Figure~\ref{fig:mqagqaviz} gives an overview of MHA, MQA, GQA, and MLKV, Figure~\ref{fig:allkvmechs} shows a more detailed view of each mechanism, and Table~\ref{tab:cache-sizes} summarizes the theoretical KV cache sizes for each. Following the notation used in Section~\ref{background}, MLKV can be written as follows:
\begin{gather}
    q_i = xW_i^Q;\; k_{j,k} = xW_{j,k}^K;\; v_{j,k} = xW_{j,k}^V \\
    o_{i,j,k} = Attention(q_i, k_{j,k}, v_{j,k}) \\
    \begin{gathered}
        MLKV_k(x) = [o_{1,1,1};o_{i,j,k};...;o_{h,g,k}]W^O \label{mlkv} \\    
        with\; j = floor(\frac{i-1}{h/g}) + 1 \\
    \end{gathered}
    \\ L_{n,k}(x) = x + MLP(x + MLKV_k(x)) \\
    \begin{gathered}
        Layers(x) = L_{1,1} \circ L_{n,k} \circ \cdot \cdot \cdot \circ L_{l,m}(x) \\
        with\; k = floor(\frac{n-1}{l/m}) + 1
    \end{gathered}
\end{gather}

\begin{table}
\centering
    \begin{tabular}{lcc}
        \hline
        \textbf{Method}& \thead{\textbf{KV Cache Size} \\ (\# Elements)} & \thead{\textbf{Cache Size} \\ \textbf{of OPT-175B} (GB)}\\
        \hline 
         MHA & $2bslhd_k$ & 144.0\\
         MQA & $2bsld_k$  & 1.5\\
         GQA & $2bslgd_k$ & 36.0\\
         MLKV& $2bsmgd_k$ & 0.375\\
        \hline
    \end{tabular}
    \caption{Theoretical KV cache sizes by number of elements for each KV sharing method. $b$ for batch size, $s$ for sequence length, $l$ for number of layers, $m$ for number of layers with their own KV heads, $h$ for number of (query) heads, $g$ for number of KV head groups, and $d_k$ for head dimension. Note that $m<l$ and $g<h$. Following the example in \ref{cache}, column 3 shows calculated cache size of OPT-175B if KV shared, given $b=8$, $s=1024$, $g=24$ for GQA and $g=1$, $m=24$ for MLKV, in float16 precision.}
    \label{tab:cache-sizes}
\end{table}

\section{Experiments}
\begin{table*}[ht]
    \centering
    \begin{tabular}{lccccccc} \toprule
         \textbf{Model Name} & \textbf{$l$} & \textbf{$h$} & \textbf{$m$} & \textbf{$g$} & \textbf{\thead{Total KV \\ Heads ($mg$)}} & \textbf{Num. Params} & \textbf{\thead{Uptrain \\ Loss}}\\ \midrule 
         Pythia-160M (baseline) & 12 & 12 & 12 & 12 & 144 & 162,322,944 & - \\ 
         Pythia-160M-GQA-48 & 12 & 12 & 12 & 4 & 48 & 162,316,800 & 2.7082 \\ 
         Pythia-160M-MLKV-48 & 12 & 12 & 4 & 12 & 48 & 162,316,800 & 2.7656 \\ 
         Pythia-160M-MQA-12 & 12 & 12 & 12 & 1 & 12 & 162,332,940 & 2.7505\\ 
         Pythia-160M-MLKV-12 & 12 & 12 & 4 & 3 & 12 & 162,332,940 & 2.8014\\ 
         Pythia-160M-MLKV-6 & 12 & 12 & 6 & 1 & 6 & 162,332,556 & 2.8013\\ 
         Pythia-160M-MLKV-4 & 12 & 12 & 4 & 1 & 4 & 162,320,132 & 2.8261\\ 
         Pythia-160M-MLKV-2 & 12 & 12 & 2 & 1 & 2 & 162,326,152 & 2.8942 \\ 
         Pythia-160M-MLKV-1 & 12 & 12 & 1 & 1 & 1 & 162,319,940 & 3.3934\\ \bottomrule
    \end{tabular}
    \caption{\label{tab:model-variants} Configurations and total KV head counts of each model variant for the experiments. $l, h, m, g$ are number of layers, number of (query) heads in a layer, number of layers with their own KV heads, and number of KV head groups in a layer, respectively. Variance in number of parameters is caused by the intermediate MLP layer sizes being only able to be increased by multiples of the model dimension. Uptrain loss is the final recorded loss of each uptraining run.}
\end{table*}

\subsection{Setup}
We utilize the Pythia suite \citep{biderman2023pythia} for our experiments, as it is open source and it provides a wide array of model sizes, even as low as 70M parameters. More specifically, we use the Pythia-160M model trained on a deduplicated The Pile dataset (pythia-160m-deduped) as our baseline. Pythia models use the same architecture as GPT-NeoX \citep{black2022gptneox}. We modify the model definition to accommodate for KV sharing, i.e. MQA, GQA, and MLKV. We follow the same data, benchmarks, and hyperparameters as in the Pythia paper. All uptraining runs are done on 2x NVIDIA A100-SXM4-80GB GPUs. Meanwhile the test runs, both for the benchmarks and inference metrics, are done on 1x NVIDIA RTX 3060 12GB.

\subsection{Models}
To see how MLKV performs at different KV head numbers and how it compares to the other KV sharing methods, we uptrain 8 variants from the baseline Pythia-160M. The 9 models are detailed in Table~\ref{tab:model-variants}. To obtain these variants, we convert the baseline model using a script that merges the KV head weights. We follow the findings in the GQA paper \citep{ainslie2023gqa} which suggests averaging the KV head weights as the best method for merging. KV heads in the same group of the same layer, as well as the heads from subsequent layers which do not have KV heads of their own, are merged by averaging. Crucially, for a fair comparison, we also make sure that each variant has the same number of total parameters. After merging, naturally there are less parameters. We compensate for this by upsizing the intermediate layer of the MLPs in each layer.

\subsection{Data}
A deduplicated version of The Pile dataset was used to train this particular Pythia model. For uptraining, ideally a portion of the same data is used. Following the results of the GQA paper \citep{ainslie2023gqa}, only 5\% is needed. Deduplicated, The Pile contains 134 million documents. For these experiments, 6 million documents are used. Additionally, we pack the data for efficiency. This means that each row is filled to the maximum sequence length (2048) with documents. To optimally do this, short documents are first packed with each other, then the long documents are truncated as needed. After packing, the data becomes 2.46 million rows of packed documents. We ensure that all uptraining runs observe the same data in the same order.

\begin{table*}[ht]
    \centering
    \begin{tabular}{lccccc}
        \toprule
        \textbf{Model} &  \textbf{ARC-e} &  \textbf{LAMBADA} &  \textbf{PIQA} &  \textbf{SciQ} &  \textbf{Average} \\
        \midrule
         Pythia-160M &     43.94 &           33.63 & 61.37 &  72.2 &        52.79 \\
        \Xhline{2\arrayrulewidth}
         Pythia-160M-GQA-48 &     41.92 &           \textbf{29.38} & 60.77 &  68.6 &        \textbf{50.17} \\
        Pythia-160M-MLKV-48 &     \textbf{42.13} &           26.18 & 59.96 &  68.9 &        49.29 \\
         Pythia-160M-MQA-12 &     40.19 &           26.74 & \textbf{61.10} &  69.7 &        49.43 \\
        Pythia-160M-MLKV-12 &     41.08 &           23.44 & 60.28 &  \textbf{70.3} &        48.78 \\
         Pythia-160M-MLKV-6 &     41.41 &           24.35 & 60.55 &  69.9 &        49.06 \\
         Pythia-160M-MLKV-4 &     40.03 &           23.64 & 60.17 &  65.4 &        47.31 \\
         Pythia-160M-MLKV-2 &     40.91 &           22.03 & 59.47 &  64.7 &        46.78 \\
         Pythia-160M-MLKV-1 &     38.26 &            8.56 & 59.25 &  58.4 &        41.12 \\
        \bottomrule
    \end{tabular}
    \caption{\label{tab:benchmark-results} Benchmarking results of all model variants. All benchmarks report accuracy and \textbf{bold} denotes the highest accuracy excluding the baseline Pythia-160M model.}
\end{table*}

\subsection{Uptraining}
The "uptraining" scheme  proposed in the GQA paper \citep{ainslie2023gqa} is used to adapt existing model checkpoints to a newly implemented KV sharing scheme by continuing pre-training. They also use the same hyperparameters from the pre-training stage. We adopt this strategy. After converting the base model weights and preparing 5\% of the original dataset, we continue training Pythia-160M using the same hyperparameters as mentioned in the paper \citep{biderman2023pythia} except for batch size and GPU count. We use a learning rate of $6 \times 10^{-4}$ with a cosine schedule and a warm-up ratio of $0.2$. We use the AdamW optimizer with $\beta_1=0.9, \beta_2=0.95, \epsilon=1 \times 10^{-8}$ and a weight decay of $0.01$. The per device batch size is set to 12, which on the 2 GPUs means a global batch size of 24. Because all model variants have around the same number of parameters, the total uptraining FLOPs are nearly equivalent, with the same runtime at around 22 hours of uptraining for all model variants. The final recorded loss value of each uptraining run is shown in Table~\ref{tab:model-variants}.

\subsection{Evaluation Method}
EleutherAI's LM Evaluation Harness \citep{eval-harness} is used as the benchmarking platform since it is convenient and was used by the Pythia paper \citep{biderman2023pythia} too. Some modifications are done to load the custom model definition. We also use the same benchmarks that were reported in the Pythia paper, but remove the ones that received near random accuracy. There are a total of 4 benchmarks used. We evaluate on the easy set of the AI2 Reasoning Challenge (ARC-e) \citep{clark2018arc}, LAMBADA \citep{paperno-etal-2016-lambada}, specifically the OpenAI variant, PIQA \citep{bisk2019piqa}, and SciQ \citep{welbl2017sciq}. All 4 benchmarks report accuracy.

Aside from benchmark performance, inference time metrics are also measured for each model. Specifically, we evaluate memory usage and throughput. This is done by a script that loads and runs the models up to their limits and measures the metrics at the same time. A dummy KV cache with length 2000 is initialized, then the model is autoregressively run token by token for 48 tokens, up to the sequence length limit. This process is timed to obtain tokens/second. Right after exiting the forward pass, memory usage is measured via the NVIDIA System Management Interface (SMI). The output is the VRAM usage measured in megabytes. We deduct the background and model memory usage from this reading. Note that due to this, the measurement results might seem to go out-of-memory prematurely, because the memory taken for tensor computations are not considered.

\begin{figure*}[t]
    \includegraphics[width=\linewidth]{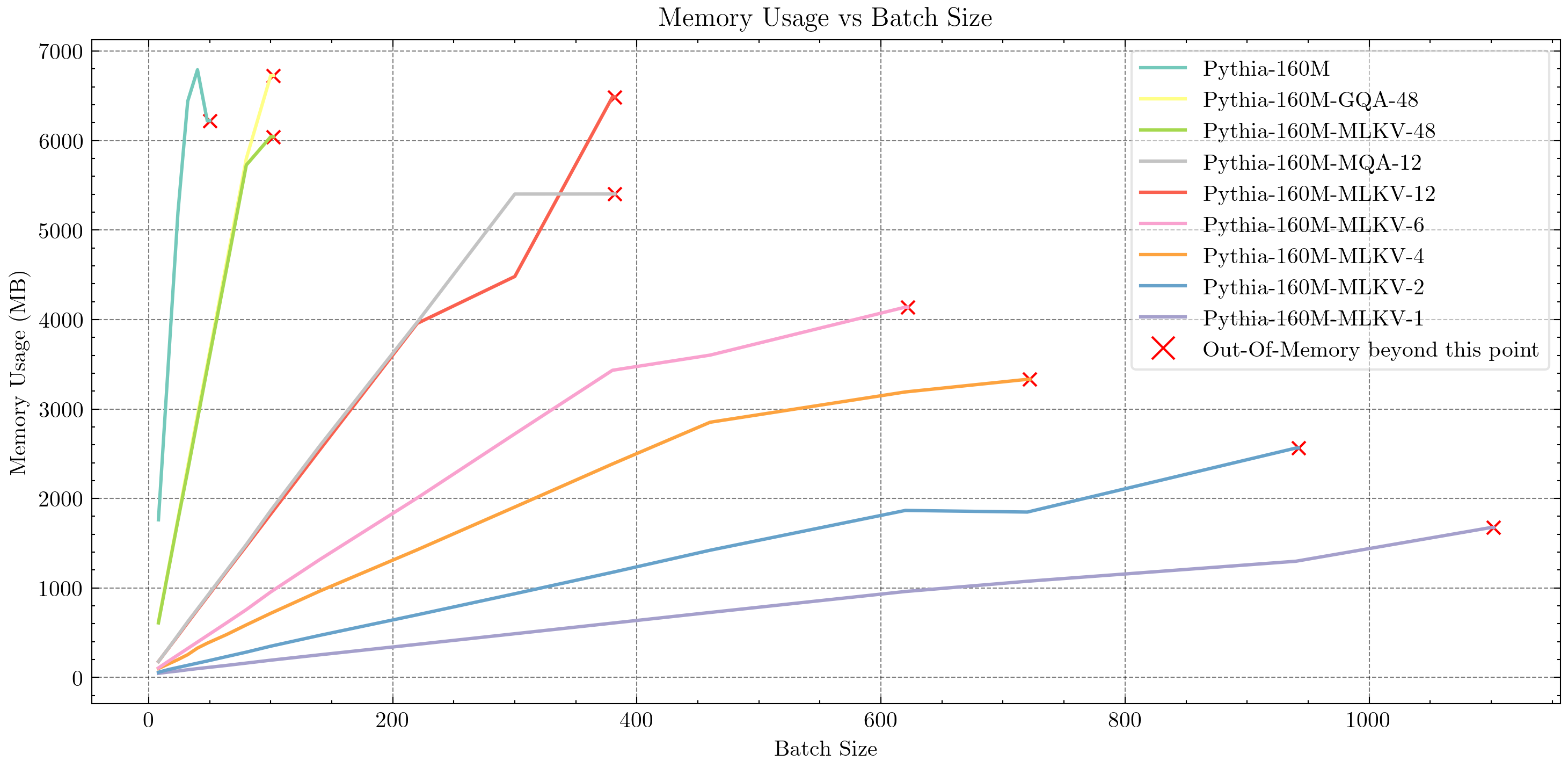}
    \caption{Line plots to visualize the inference time memory measurements in terms of the batch sizes that can be achieved by each model. The red 'X' indicates that beyond that batch size, an out-of-memory error will occur.}
    \label{fig:memvsbatch}
\end{figure*}

\begin{figure}[t]
    \centering
    \includegraphics[width=\columnwidth]{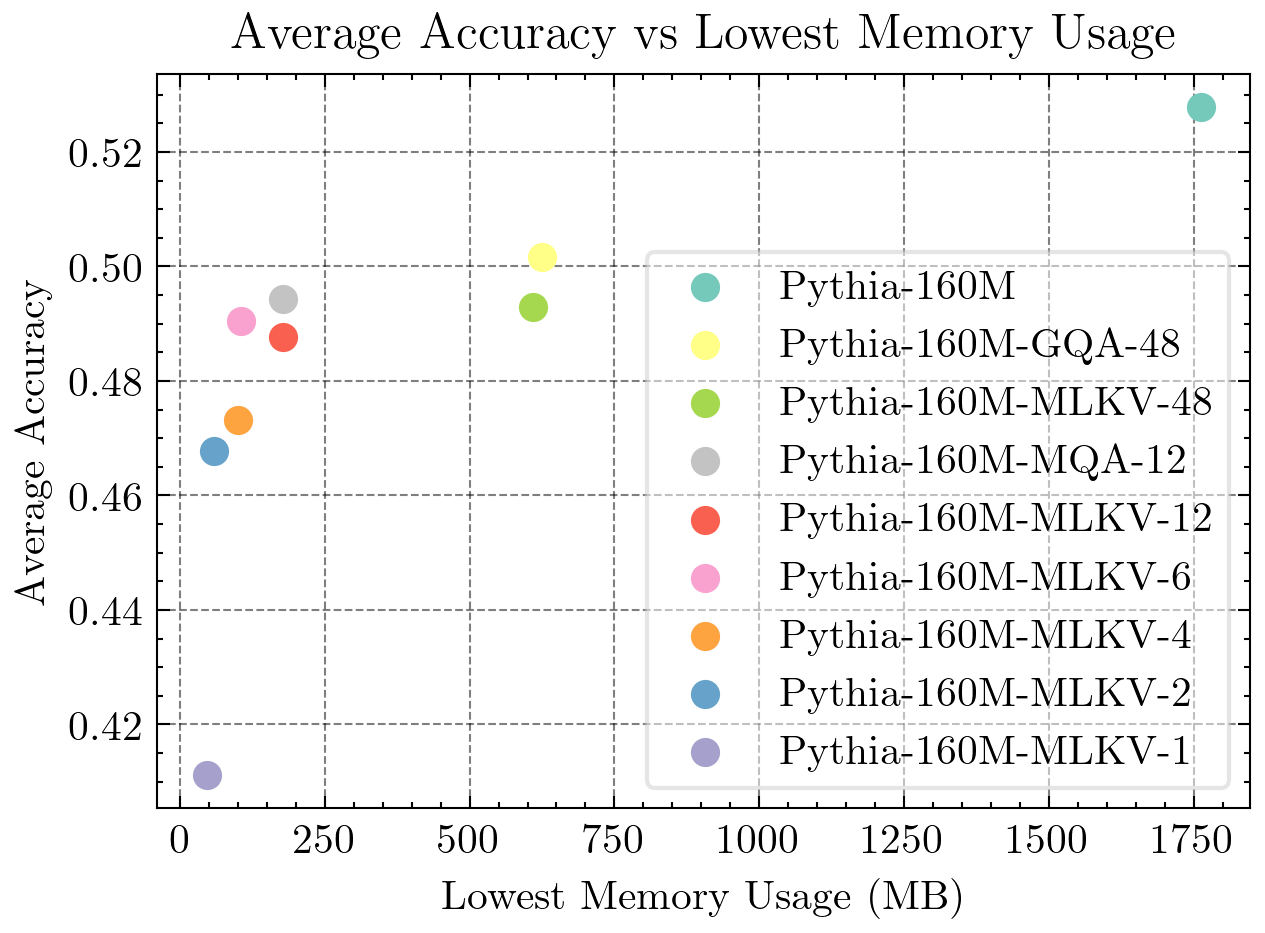}
    \caption{Average accuracy vs lowest recorded memory usage (this is at a minimum batch size but memory scales the same way as it increases). Pareto optimality resides in the left upper corner of the plot.}
    \label{fig:accvsmem}
\end{figure}

\section{Results}
\subsection{Benchmarks}
Table~\ref{tab:benchmark-results} contains the results of each model variant on the benchmarks. Firstly, for the models that compare between GQA/MQA and MLKV, MLKV mostly underperforms at the same KV head count. At 48 heads, GQA-48 performs better in all benchmarks but SciQ compared to MLKV-48. Meanwhile at 12 heads, MLKV-12 does better in ARC-e but is worse in others compared to MQA-12. This is also reflected in the average accuracy.

Overall, there is a clear trend of decreasing accuracy as the KV head count goes down. At the most extreme, MLKV-1 significantly degrades the model, rendering it basically unusable, as indicated by LAMBADA. However, other head counts show much more promising results. Looking at the average accuracy, KV head counts at 48, 12, and 6 are very close to each other, indicating that it is still worth it to cut down to a head count lower than the layer count. Accuracy goes down more noticeably for MLKV-4 and MLKV-2, but not by a drastic amount. These head counts can still be considered for the accuracy/memory trade-off it provides.

\begin{table*}[ht]
    \centering
    \begin{tabular}{lccccc}
        \toprule
        \textbf{Model} &  \textbf{ARC-e} &  \textbf{LAMBADA} &  \textbf{PIQA} &  \textbf{SciQ} &  \textbf{Average} \\
        \midrule
         Pythia-410M &      51.6 &           51.93 & 67.19 &  82.7 &        63.36 \\
        \Xhline{2\arrayrulewidth}
 Pythia-410M-MQA-24 &     \textbf{46.04} &           \textbf{34.06} & \textbf{63.82} &  75.1 &        \textbf{54.76} \\
Pythia-410M-MLKV-12 &     44.70 &           33.11 & 62.89 &  \textbf{75.4} &        54.02 \\
 Pythia-410M-MLKV-8 &     \textbf{46.04} &           31.28 & 62.62 &  72.9 &        53.21 \\
        \bottomrule
    \end{tabular}
    \caption{\label{tab:410m-benchmark-results} Benchmarking results of the Pythia-410M model variants. All benchmarks report accuracy and \textbf{bold} denotes the highest accuracy excluding the baseline Pythia-410M model.}
\end{table*}

\subsection{Inference Time Measurements}

We evaluate memory usage while generating various batch sizes up to each model's limit, as can be seen in Figure~\ref{fig:memvsbatch}. Most obvious is the way each KV head count scales in memory as batch size increases. They scale linearly at rate that matches each KV head count. This is expected given the theoretical KV cache tensor sizes shown in Tabel~\ref{tab:cache-sizes}. The visual clearly shows how significant the memory benefits are as KV is shared. The incline of each line plot determines the maximum batch size possible of each model variant. The baseline can only go up to 48 on our setup, meanwhile at the most extreme, MLKV-2 and MLKV-1 go up to 940 and 1100 respectively. Importantly though, these measurements would also apply for increasing sequence length, as it scales in the same way as batch size. If we set a fixed batch size, and replace the X axis with sequence length, it would generate the exact same plots.

We also plot the average accuracy reported from benchmarks and the memory usage of all model variants in Figure~\ref{fig:accvsmem}. Ideally, we want high accuracy with low memory usage. The variants GQA-48, MLKV-48, and MLKV-12 show logical departures of both accuracy and memory compared to baseline. However, it can be seen that the most Pareto optimal models are MQA-12 and MLKV-6, both being in the upper left side of the curve. These variants provide the most favorable trade-off in accuracy/memory, according to the data. Accuracy goes down a small amount as we go to MLKV-4 and MLKV-2, but they do not degrade the same way as MLKV-1 and still display a reasonable exchange between accuracy and memory benefits.

\begin{figure}[t]
    \centering
    \includegraphics[width=\columnwidth]{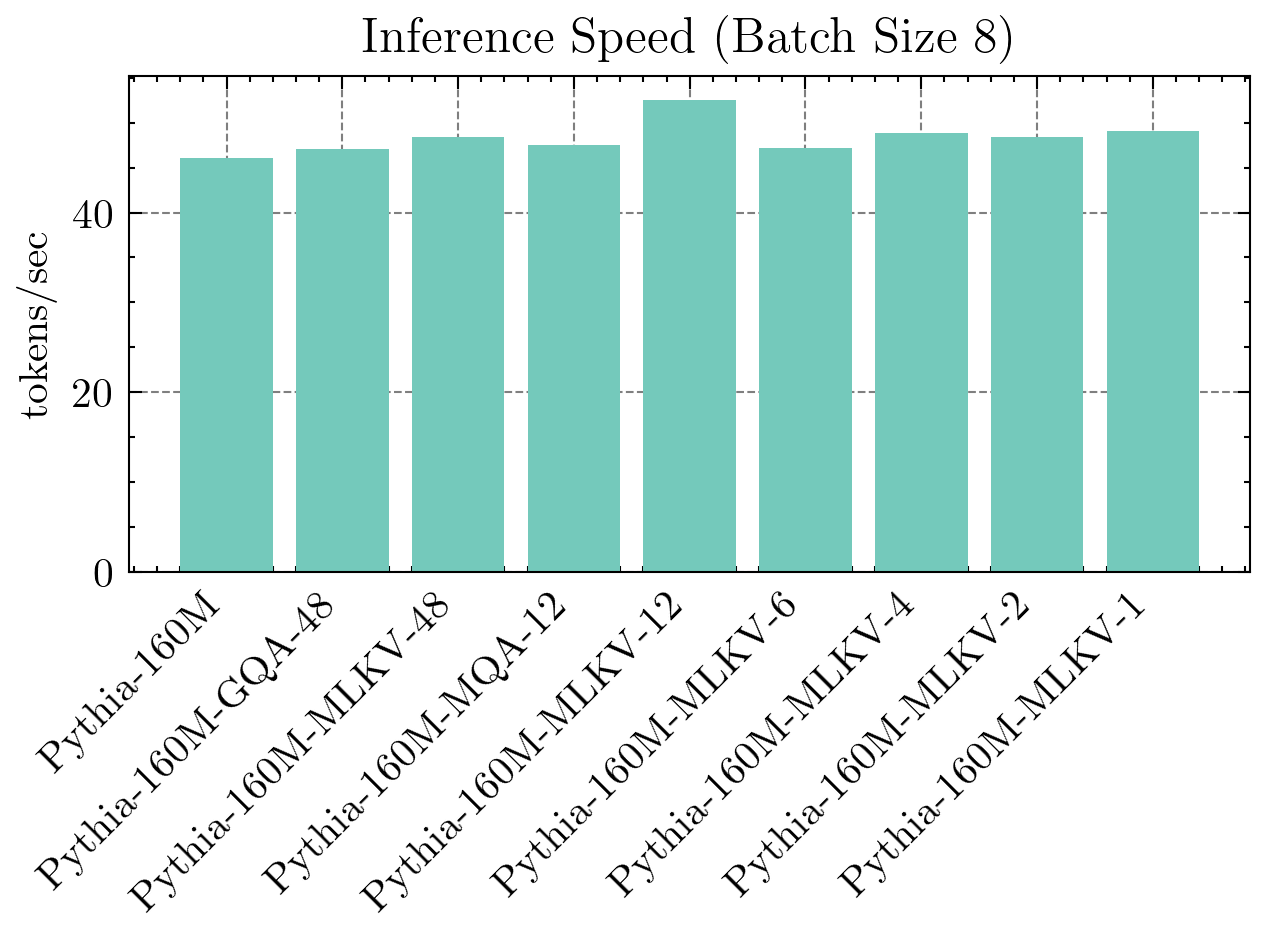}
    \caption{Measured throughput of each model in tokens/second, recorded at a batch size of 8.}
    \label{fig:speed}
\end{figure}

Lastly, the measured throughput of each model at batch size 8 is shown in Figure~\ref{fig:speed}. Emperically, we do not see any significant speed-up through MLKV. Theoretically, fetching KV cache at every layer requires the same overhead, may it be a shared KV head or not. This might be improved through a more custom, optimized implementation. Our implementation is generalized to accommodate for GQA/MQA configurations.

\begin{figure}[t]
    \centering
    \includegraphics[width=\columnwidth]{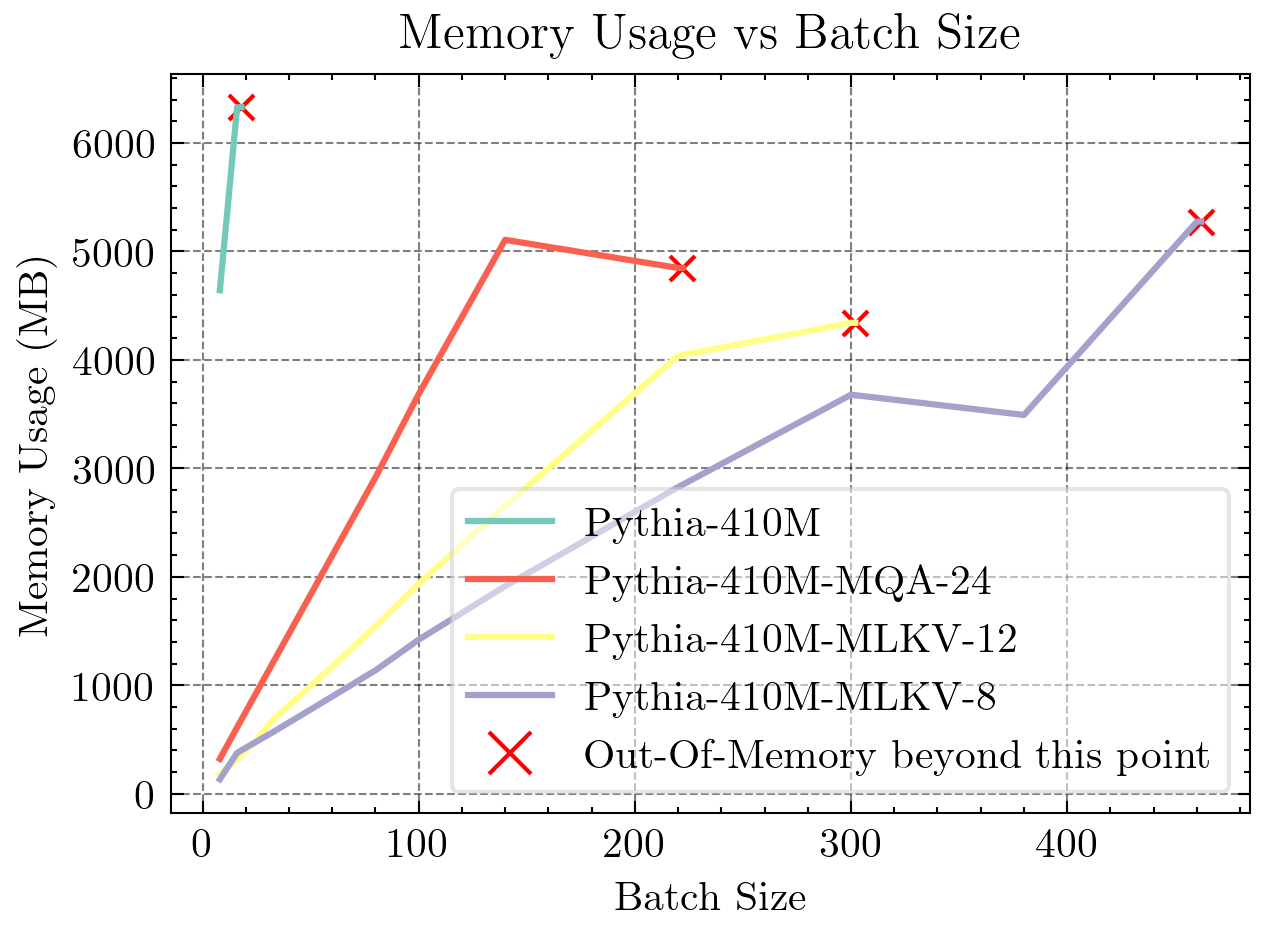}
    \caption{Inference time memory measurements of the 410M parameter models. The red 'X' indicates that beyond that batch size, an OOM error will occur.}
    \label{fig:accvsmem410m}
\end{figure}

\subsection{Scaling Up}
The experiments done above were done on a small scale for ablation to see what KV head counts work for optimal memory-accuracy trade-off. However, it is also important to verify whether or not these gains and losses are also applicable at a larger scale. For that, we do further uptraining and evaluation of a larger model, Pythia-410M. Specifically, since we have seen the most optimal results from configurations with $mg=l/2$ and even down to $mg=l/3$, we uptrain only these variants and also an MQA variant to compare with baseline. Since Pythia-410M has a total of 24 layers ($l=24$) and 12 query heads in each layer ($h=12$), we uptrain an MQA variant with $m=24,g=1$ resulting in Pythia-410M-MQA-24, and two MLKV variants with $m=12,g=1$ and $m=8,g=1$ resulting in Pythia-410M-MLKV-12 and Pythia-410M-MLKV-8 respectively. Results are shown in Table~\ref{tab:410m-benchmark-results} and Figure~\ref{fig:accvsmem410m}.

\section{Discussion}
The experiments show a clear trade-off between memory and accuracy. It is left to architecture designers to choose how much to sacrifice for the memory benefits, with multiple things to consider. For KV head counts above or equal to the number of layers, i.e. $mg \geq l$, given the performance shown, it is still better to use GQA/MQA instead of MLKV. We theorize that this is because having KV heads in multiple layers is more important than having multiple KV heads in the same layer. In other words, designers should sacrifice KV heads in-layer first (via GQA/MQA) and cross-layer second (via MLKV). For use cases with tighter memory requirements that require $mg<l$, then MLKV is the only way. We show that this design decision is still very much viable. We find that $mg=l/2$ with MLKV performs very near MQA in all scales tested, which means it should be a relatively easy decision if a KV cache is needed that is half the size of what is provided by MQA. For requirements below that, we find $mg=l/3$ and even $mg=l/6$ to still be usable, without drastic degradation. Anything below that becomes questionable. It is clear that $mg=1$ is too extreme and results in model collapse. The transformer benefits from the multiple recomputations of key-values from layer-to-layer, but can still compromise some level of it for the benefit of memory.

\section{Related Work}
KV cache optimization methods on dimensions other than $h$ and $l$ have also been done. Approaches in reducing the sequence length $s$ of the cache do it by compressing information in the context. Attention mask variations like sliding windows \citep{zaheer2020bigbird}, dilated sliding windows \citep{beltagy2020longformer}, and attention sinks \citep{xiao2023attentionsink} attempt to limit the receptive field of tokens to some length smaller than the actual sequence length, with the assumption that the information of the tokens before will be compressed to some other positions. A more concrete compression-based solution is SCISSORHANDS \citep{liu2023scissorhands} which only keeps pivotal tokens with the assumption of \textit{persistence of importance}, such that the KV cache length can be kept short. FastGen \citep{ge2024fastgen} employs an adaptive KV cache compression strategy based on specific policies on special tokens, punctuation, locality, and frequency of different positions to determine which KV heads to prune.

Further improvements through low-rank compression were introduced by Multi-Latent Attention (MLA) with DeepSeek-V2 \citep{deepseekai2024deepseekv2}. All key-values are projected down into smaller latent vectors in each layer, which are cached with a much smaller memory footprint. These vectors are then projected up into all the needed key-values for all heads in the attention mechanism. Furthermore, You-Only-Cache-Once (YOCO) \citep{sun2024cacheoncedecoderdecoderarchitectures} uses the first half of the layers to create an intermediate embedding, that is then projected into semi-global keys and values which are used for all layers of the latter half of the transformer, basically removing half of the KV cache layers. 

Moreover, we are aware of the contemporaneous work on Cross-Layer Attention (CLA) \citep{brandon2024reducing}, published when we were nearing the end of finishing our work on MLKV. CLA was experimented on larger models trained from scratch, instead of the uptraining practice in our work, which utilizes available model checkpoints. Additional ablations were done with CLA, including non-uniform layer sharing patterns, but missing more extreme configurations like MLKV-2 and MLKV-1. The ablations are also based on a KV cache memory budget equivalence, instead of matching parameter counts in MLKV.

\section{Conclusion}
Our proposed KV sharing method Multi-Layer Key-Value (MLKV) provides the option to further reduce KV cache size in transformers beyond what was possible with GQA and MQA. By sharing KV heads not only inside a layer but also between layers, we can reduce the total KV head count to lower than the number of layers in the transformer. We show through experiments that reductions of a factor up to 6x in cache size compared to MQA are possible and provide a fair accuracy/memory trade-off. We recommend sharing to every second layer (KV head count equal to half the number of layers) for 2x reduction from MQA with very minimal reduction in accuracy, but ultimately give the option to architecture designers to decide if even lower number of KV heads is needed for more memory constrained use cases.

\section*{Limitations}
We only evaluated MLKV on decoder-only models, while encoder-decoder transformers could also benefit from this KV sharing method on their decoders. Additionally, the scale of our experiments was relatively small, conducted on models with 160 million and 410 million parameters. However, models at the billion-parameter scale are becoming much more common, and MLKV has yet to be tested at such scales. Due to the relatively small size of the model we tested, the number and variety of downstream tasks that can be reliably used to compare performance is limited. Therefore, the impact of MLKV on other tasks remains to be seen. Furthermore, we did not train models from scratch and thus do not know how MLKV performs on models that are natively imbued with this KV sharing scheme from the pre-training phase.


\bibliography{custom}

\begin{thebibliography}{21}
\expandafter\ifx\csname natexlab\endcsname\relax\def\natexlab#1{#1}\fi

\bibitem[{Ainslie et~al.(2023)Ainslie, Lee-Thorp, de~Jong, Zemlyanskiy, Lebrón, and Sanghai}]{ainslie2023gqa}
Joshua Ainslie, James Lee-Thorp, Michiel de~Jong, Yury Zemlyanskiy, Federico Lebrón, and Sumit Sanghai. 2023.
\newblock \href {http://arxiv.org/abs/2305.13245} {Gqa: Training generalized multi-query transformer models from multi-head checkpoints}.

\bibitem[{Beltagy et~al.(2020)Beltagy, Peters, and Cohan}]{beltagy2020longformer}
Iz~Beltagy, Matthew~E. Peters, and Arman Cohan. 2020.
\newblock \href {http://arxiv.org/abs/2004.05150} {Longformer: The long-document transformer}.
\newblock \emph{CoRR}, abs/2004.05150.

\bibitem[{Biderman et~al.(2023)Biderman, Schoelkopf, Anthony, Bradley, O'Brien, Hallahan, Khan, Purohit, Prashanth, Raff, Skowron, Sutawika, and van~der Wal}]{biderman2023pythia}
Stella Biderman, Hailey Schoelkopf, Quentin Anthony, Herbie Bradley, Kyle O'Brien, Eric Hallahan, Mohammad~Aflah Khan, Shivanshu Purohit, USVSN~Sai Prashanth, Edward Raff, Aviya Skowron, Lintang Sutawika, and Oskar van~der Wal. 2023.
\newblock \href {http://arxiv.org/abs/2304.01373} {Pythia: A suite for analyzing large language models across training and scaling}.

\bibitem[{Bisk et~al.(2019)Bisk, Zellers, Bras, Gao, and Choi}]{bisk2019piqa}
Yonatan Bisk, Rowan Zellers, Ronan~Le Bras, Jianfeng Gao, and Yejin Choi. 2019.
\newblock \href {http://arxiv.org/abs/1911.11641} {Piqa: Reasoning about physical commonsense in natural language}.

\bibitem[{Black et~al.(2022)Black, Biderman, Hallahan, Anthony, Gao, Golding, He, Leahy, McDonell, Phang, Pieler, Prashanth, Purohit, Reynolds, Tow, Wang, and Weinbach}]{black2022gptneox}
Sidney Black, Stella Biderman, Eric Hallahan, Quentin Anthony, Leo Gao, Laurence Golding, Horace He, Connor Leahy, Kyle McDonell, Jason Phang, Michael Pieler, Usvsn~Sai Prashanth, Shivanshu Purohit, Laria Reynolds, Jonathan Tow, Ben Wang, and Samuel Weinbach. 2022.
\newblock \href {https://doi.org/10.18653/v1/2022.bigscience-1.9} {{GPT}-{N}eo{X}-20{B}: An open-source autoregressive language model}.
\newblock In \emph{Proceedings of BigScience Episode {\#}5 -- Workshop on Challenges {\&} Perspectives in Creating Large Language Models}, pages 95--136, virtual+Dublin. Association for Computational Linguistics.

\bibitem[{Brandon et~al.(2024)Brandon, Mishra, Nrusimha, Panda, and Kelly}]{brandon2024reducing}
William Brandon, Mayank Mishra, Aniruddha Nrusimha, Rameswar Panda, and Jonathan~Ragan Kelly. 2024.
\newblock \href {http://arxiv.org/abs/2405.12981} {Reducing transformer key-value cache size with cross-layer attention}.

\bibitem[{Clark et~al.(2018)Clark, Cowhey, Etzioni, Khot, Sabharwal, Schoenick, and Tafjord}]{clark2018arc}
Peter Clark, Isaac Cowhey, Oren Etzioni, Tushar Khot, Ashish Sabharwal, Carissa Schoenick, and Oyvind Tafjord. 2018.
\newblock \href {http://arxiv.org/abs/1803.05457} {Think you have solved question answering? try arc, the ai2 reasoning challenge}.

\bibitem[{DeepSeek-AI et~al.(2024)DeepSeek-AI, Liu, Feng, Wang, Wang, Liu, Zhao, Dengr, Ruan, Dai, Guo, Yang, Chen, Ji, Li, Lin, Luo, Hao, Chen, Li, Zhang, Xu, Yang, Zhang, Ding, Xin, Gao, Li, Qu, Cai, Liang, Guo, Ni, Li, Chen, Yuan, Qiu, Song, Dong, Gao, Guan, Wang, Zhang, Xu, Xia, Zhao, Zhang, Li, Wang, Zhang, Zhang, Tang, Li, Tian, Huang, Wang, Zhang, Zhu, Chen, Du, Chen, Jin, Ge, Pan, Xu, Chen, Li, Lu, Zhou, Chen, Wu, Ye, Ma, Wang, Zhou, Yu, Zhou, Zheng, Wang, Pei, Yuan, Sun, Xiao, Zeng, An, Liu, Liang, Gao, Zhang, Li, Jin, Wang, Bi, Liu, Wang, Shen, Chen, Chen, Nie, Sun, Wang, Liu, Xie, Yu, Song, Zhou, Yang, Lu, Su, Wu, Li, Wei, Zhu, Xu, Huang, Li, Zhao, Sun, Li, Wang, Zheng, Zhang, Xiong, Zhao, He, Tang, Piao, Dong, Tan, Liu, Wang, Guo, Zhu, Wang, Zou, Zha, Ma, Yan, You, Liu, Ren, Ren, Sha, Fu, Huang, Zhang, Xie, Hao, Shao, Wen, Xu, Zhang, Li, Wang, Gu, Li, and Xie}]{deepseekai2024deepseekv2}
DeepSeek-AI, Aixin Liu, Bei Feng, Bin Wang, Bingxuan Wang, Bo~Liu, Chenggang Zhao, Chengqi Dengr, Chong Ruan, Damai Dai, Daya Guo, Dejian Yang, Deli Chen, Dongjie Ji, Erhang Li, Fangyun Lin, Fuli Luo, Guangbo Hao, Guanting Chen, Guowei Li, H.~Zhang, Hanwei Xu, Hao Yang, Haowei Zhang, Honghui Ding, Huajian Xin, Huazuo Gao, Hui Li, Hui Qu, J.~L. Cai, Jian Liang, Jianzhong Guo, Jiaqi Ni, Jiashi Li, Jin Chen, Jingyang Yuan, Junjie Qiu, Junxiao Song, Kai Dong, Kaige Gao, Kang Guan, Lean Wang, Lecong Zhang, Lei Xu, Leyi Xia, Liang Zhao, Liyue Zhang, Meng Li, Miaojun Wang, Mingchuan Zhang, Minghua Zhang, Minghui Tang, Mingming Li, Ning Tian, Panpan Huang, Peiyi Wang, Peng Zhang, Qihao Zhu, Qinyu Chen, Qiushi Du, R.~J. Chen, R.~L. Jin, Ruiqi Ge, Ruizhe Pan, Runxin Xu, Ruyi Chen, S.~S. Li, Shanghao Lu, Shangyan Zhou, Shanhuang Chen, Shaoqing Wu, Shengfeng Ye, Shirong Ma, Shiyu Wang, Shuang Zhou, Shuiping Yu, Shunfeng Zhou, Size Zheng, T.~Wang, Tian Pei, Tian Yuan, Tianyu Sun, W.~L. Xiao, Wangding Zeng, Wei An, Wen
  Liu, Wenfeng Liang, Wenjun Gao, Wentao Zhang, X.~Q. Li, Xiangyue Jin, Xianzu Wang, Xiao Bi, Xiaodong Liu, Xiaohan Wang, Xiaojin Shen, Xiaokang Chen, Xiaosha Chen, Xiaotao Nie, Xiaowen Sun, Xiaoxiang Wang, Xin Liu, Xin Xie, Xingkai Yu, Xinnan Song, Xinyi Zhou, Xinyu Yang, Xuan Lu, Xuecheng Su, Y.~Wu, Y.~K. Li, Y.~X. Wei, Y.~X. Zhu, Yanhong Xu, Yanping Huang, Yao Li, Yao Zhao, Yaofeng Sun, Yaohui Li, Yaohui Wang, Yi~Zheng, Yichao Zhang, Yiliang Xiong, Yilong Zhao, Ying He, Ying Tang, Yishi Piao, Yixin Dong, Yixuan Tan, Yiyuan Liu, Yongji Wang, Yongqiang Guo, Yuchen Zhu, Yuduan Wang, Yuheng Zou, Yukun Zha, Yunxian Ma, Yuting Yan, Yuxiang You, Yuxuan Liu, Z.~Z. Ren, Zehui Ren, Zhangli Sha, Zhe Fu, Zhen Huang, Zhen Zhang, Zhenda Xie, Zhewen Hao, Zhihong Shao, Zhiniu Wen, Zhipeng Xu, Zhongyu Zhang, Zhuoshu Li, Zihan Wang, Zihui Gu, Zilin Li, and Ziwei Xie. 2024.
\newblock \href {http://arxiv.org/abs/2405.04434} {Deepseek-v2: A strong, economical, and efficient mixture-of-experts language model}.

\bibitem[{Gao et~al.(2023)Gao, Tow, Abbasi, Biderman, Black, DiPofi, Foster, Golding, Hsu, Le~Noac'h, Li, McDonell, Muennighoff, Ociepa, Phang, Reynolds, Schoelkopf, Skowron, Sutawika, Tang, Thite, Wang, Wang, and Zou}]{eval-harness}
Leo Gao, Jonathan Tow, Baber Abbasi, Stella Biderman, Sid Black, Anthony DiPofi, Charles Foster, Laurence Golding, Jeffrey Hsu, Alain Le~Noac'h, Haonan Li, Kyle McDonell, Niklas Muennighoff, Chris Ociepa, Jason Phang, Laria Reynolds, Hailey Schoelkopf, Aviya Skowron, Lintang Sutawika, Eric Tang, Anish Thite, Ben Wang, Kevin Wang, and Andy Zou. 2023.
\newblock \href {https://doi.org/10.5281/zenodo.10256836} {A framework for few-shot language model evaluation}.

\bibitem[{Ge et~al.(2024)Ge, Zhang, Liu, Zhang, Han, and Gao}]{ge2024fastgen}
Suyu Ge, Yunan Zhang, Liyuan Liu, Minjia Zhang, Jiawei Han, and Jianfeng Gao. 2024.
\newblock \href {http://arxiv.org/abs/2310.01801} {Model tells you what to discard: Adaptive kv cache compression for llms}.

\bibitem[{Geva et~al.(2021)Geva, Schuster, Berant, and Levy}]{geva-etal-2021-transformer}
Mor Geva, Roei Schuster, Jonathan Berant, and Omer Levy. 2021.
\newblock \href {https://doi.org/10.18653/v1/2021.emnlp-main.446} {Transformer feed-forward layers are key-value memories}.
\newblock In \emph{Proceedings of the 2021 Conference on Empirical Methods in Natural Language Processing}, pages 5484--5495, Online and Punta Cana, Dominican Republic. Association for Computational Linguistics.

\bibitem[{Liu et~al.(2023)Liu, Desai, Liao, Wang, Xie, Xu, Kyrillidis, and Shrivastava}]{liu2023scissorhands}
Zichang Liu, Aditya Desai, Fangshuo Liao, Weitao Wang, Victor Xie, Zhaozhuo Xu, Anastasios Kyrillidis, and Anshumali Shrivastava. 2023.
\newblock \href {http://arxiv.org/abs/2305.17118} {Scissorhands: Exploiting the persistence of importance hypothesis for llm kv cache compression at test time}.

\bibitem[{Paperno et~al.(2016)Paperno, Kruszewski, Lazaridou, Pham, Bernardi, Pezzelle, Baroni, Boleda, and Fern{\'a}ndez}]{paperno-etal-2016-lambada}
Denis Paperno, Germ{\'a}n Kruszewski, Angeliki Lazaridou, Ngoc~Quan Pham, Raffaella Bernardi, Sandro Pezzelle, Marco Baroni, Gemma Boleda, and Raquel Fern{\'a}ndez. 2016.
\newblock \href {https://doi.org/10.18653/v1/P16-1144} {The {LAMBADA} dataset: Word prediction requiring a broad discourse context}.
\newblock In \emph{Proceedings of the 54th Annual Meeting of the Association for Computational Linguistics (Volume 1: Long Papers)}, pages 1525--1534, Berlin, Germany. Association for Computational Linguistics.

\bibitem[{Pope et~al.(2022)Pope, Douglas, Chowdhery, Devlin, Bradbury, Levskaya, Heek, Xiao, Agrawal, and Dean}]{pope2022efficiently}
Reiner Pope, Sholto Douglas, Aakanksha Chowdhery, Jacob Devlin, James Bradbury, Anselm Levskaya, Jonathan Heek, Kefan Xiao, Shivani Agrawal, and Jeff Dean. 2022.
\newblock \href {http://arxiv.org/abs/2211.05102} {Efficiently scaling transformer inference}.

\bibitem[{Shazeer(2019)}]{shazeer2019fast}
Noam Shazeer. 2019.
\newblock \href {http://arxiv.org/abs/1911.02150} {Fast transformer decoding: One write-head is all you need}.

\bibitem[{Sun et~al.(2024)Sun, Dong, Zhu, Huang, Wang, Ma, Zhang, Wang, and Wei}]{sun2024cacheoncedecoderdecoderarchitectures}
Yutao Sun, Li~Dong, Yi~Zhu, Shaohan Huang, Wenhui Wang, Shuming Ma, Quanlu Zhang, Jianyong Wang, and Furu Wei. 2024.
\newblock \href {http://arxiv.org/abs/2405.05254} {You only cache once: Decoder-decoder architectures for language models}.

\bibitem[{Vaswani et~al.(2017)Vaswani, Shazeer, Parmar, Uszkoreit, Jones, Gomez, Kaiser, and Polosukhin}]{vaswani2017attention}
Ashish Vaswani, Noam Shazeer, Niki Parmar, Jakob Uszkoreit, Llion Jones, Aidan~N. Gomez, \L{}ukasz Kaiser, and Illia Polosukhin. 2017.
\newblock Attention is all you need.
\newblock In \emph{Proceedings of the 31st International Conference on Neural Information Processing Systems}, NIPS'17, page 6000–6010, Red Hook, NY, USA. Curran Associates Inc.

\bibitem[{Welbl et~al.(2017)Welbl, Liu, and Gardner}]{welbl2017sciq}
Johannes Welbl, Nelson~F. Liu, and Matt Gardner. 2017.
\newblock \href {http://arxiv.org/abs/1707.06209} {Crowdsourcing multiple choice science questions}.

\bibitem[{Xiao et~al.(2023)Xiao, Tian, Chen, Han, and Lewis}]{xiao2023attentionsink}
Guangxuan Xiao, Yuandong Tian, Beidi Chen, Song Han, and Mike Lewis. 2023.
\newblock \href {http://arxiv.org/abs/2309.17453} {Efficient streaming language models with attention sinks}.

\bibitem[{Zaheer et~al.(2020)Zaheer, Guruganesh, Dubey, Ainslie, Alberti, Onta{\~{n}}{\'{o}}n, Pham, Ravula, Wang, Yang, and Ahmed}]{zaheer2020bigbird}
Manzil Zaheer, Guru Guruganesh, Avinava Dubey, Joshua Ainslie, Chris Alberti, Santiago Onta{\~{n}}{\'{o}}n, Philip Pham, Anirudh Ravula, Qifan Wang, Li~Yang, and Amr Ahmed. 2020.
\newblock \href {http://arxiv.org/abs/2007.14062} {Big bird: Transformers for longer sequences}.
\newblock \emph{CoRR}, abs/2007.14062.

\bibitem[{Zhang et~al.(2022)Zhang, Roller, Goyal, Artetxe, Chen, Chen, Dewan, Diab, Li, Lin, Mihaylov, Ott, Shleifer, Shuster, Simig, Koura, Sridhar, Wang, and Zettlemoyer}]{zhang2022opt}
Susan Zhang, Stephen Roller, Naman Goyal, Mikel Artetxe, Moya Chen, Shuohui Chen, Christopher Dewan, Mona Diab, Xian Li, Xi~Victoria Lin, Todor Mihaylov, Myle Ott, Sam Shleifer, Kurt Shuster, Daniel Simig, Punit~Singh Koura, Anjali Sridhar, Tianlu Wang, and Luke Zettlemoyer. 2022.
\newblock \href {http://arxiv.org/abs/2205.01068} {Opt: Open pre-trained transformer language models}.

\end{thebibliography}




\end{document}